\documentclass[letterpaper, 10 pt, conference]{ieeeconf}  

\IEEEoverridecommandlockouts                              

\usepackage{amssymb}  
\usepackage{amsmath}
\usepackage{graphicx}
\usepackage{xcolor}
\usepackage{float}
\usepackage{booktabs}
\usepackage{diagbox}
\usepackage{makecell}
\usepackage{amsmath}
\usepackage{diagbox}
\usepackage{booktabs}   
\usepackage{tabularx}   
\usepackage{array}    
\usepackage{amsmath}    
\usepackage{amssymb}    
\usepackage{multirow}

\title{\LARGE \bf
Computational Design of Customized Tendon-Driven \\
Origami End-Effectors for Enveloping Grasp
}
\author{Zizhuo Wang, Peiwen Yang, Shuguang Li\\
Tsinghua University}

\begin{document}

\maketitle
\thispagestyle{empty}
\pagestyle{empty}

\begin{abstract}
Robotic grasping is key to daily and industrial manipulation. Object-specific grippers can exploit known geometry to improve reliability while reducing control complexity. Origami-inspired designs are well suited to these applications because their crease patterns enable predictable, programmable shape transformations. Yet mapping a target object to a deployable origami gripper remains challenging: the origami structure must fit the object, satisfy foldability and fabrication constraints, and support compact actuation. We present an integrated computational design and fabrication pipeline for customized, tendon-driven origami end-effectors for enveloping grasp. Given a target mesh, our method uses reinforcement learning in simulation to automatically design an axisymmetric enveloping gripper and generate its crease pattern. We fabricate the resulting structure using laminated sheets and integrate it with a compact single-motor tendon actuation module. Experiments on YCB objects demonstrate a $100\%$ target-object grasp success rate and quasi-static maximum holding forces of $29.50\sim130.13~\mathrm{N}$ ($16.7\times\sim83.2\times$ the gripper's weight) across the four target objects. These results validate the feasibility and effectiveness of the proposed computational design, fabrication, and actuation pipeline for customized origami grippers.

\end{abstract}



\section{Introduction}
Robotic grasping is a fundamental capability for manipulation in daily life and industrial manufacturing~\cite{long2020systematic}. 
When a known task or target object requires higher grasp reliability, customized grippers provide an effective alternative to universal end-effectors~\cite{schroeffer2019automated,gilday2025embodied,ha2021fit2form,kodnongbua2023computational,soft_enveloping_gripper,honarpardaz2017fast}. 
By embedding object and task characteristics into the end-effector, they shift part of the problem from control to morphology, improving contact efficiency, grasp stability, and payload capacity while reducing control complexity. 
However, most customized grippers remain rigid, relying on manual design and iterative trial-and-error fabrication---limiting their ability to be deployed safely, rapidly, and cost-effectively.


Soft grippers improve safety and grasping tolerance through passive compliance~\cite{rus2015design,shintake2018soft,laschi2016soft}. 
Yet compliance alone does not ensure predictable deformation or high grasping capacity.
Origami offers a complementary design paradigm. 
Its crease pattern encodes where a sheet bends and how the body transforms, making deformation programmable and predictable while retaining light weight, gentle contact, and low-cost fabrication~\cite{rus2018design,mahadevan2005self,dudte2016programming,felton2014method}. 
These properties make origami well suited to customized grippers, whose morphology must be tailored to diverse requirements.

Within this design space, morphology determines attainable grasp performance.
Many soft grippers use discrete finger-like architectures~\cite{jin2024computational,liu2024tendon}. 
These designs are simple and adaptive, but they constrain objects through sparse contacts and local finger bending. 
In contrast, an axisymmetric enveloping gripper can surround the object with a closed-loop structure, providing distributed contact and geometric enclosure to improve grasp adaptability, stability, and load capacity~\cite{soft_enveloping_gripper,seo2016theory,envelop_soft,Sui2022}. 
The vacuum-driven ``magic-ball'' origami gripper demonstrated the potential of this principle by combining a hollow origami skeleton with a flexible membrane to achieve robust enveloping grasps~\cite{magicball}. 
However, both its overall morphology and crease pattern are manually prescribed, without a systematic way to tailor the gripper structure to different requirements.

\begin{figure}[t]
    \centering
    \includegraphics[width=0.95\columnwidth]{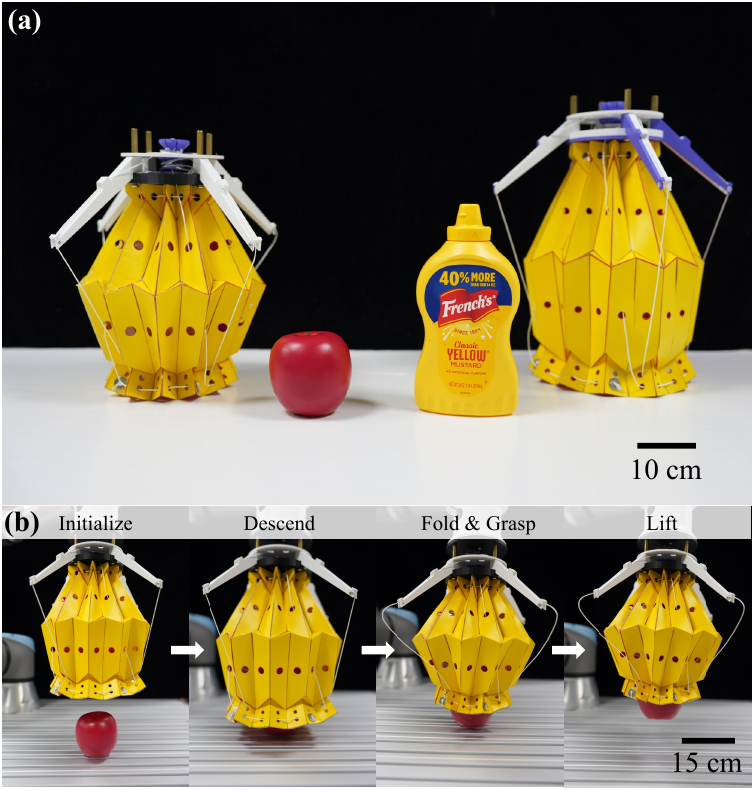}
    \caption{\textbf{Overview of our customized grippers.}
    (a) We designed customized origami grippers for different objects. They are lightweight, soft, and electrically tendon-driven, demonstrating stable and powerful grasping capabilities.
    (b) The grasping process of our gripper.}
    \label{fig:head}
\end{figure}

Computational design provides a way to move beyond manually prescribed robot morphologies. 
It has been widely used in both rigid and soft robotic systems~\cite{schroeffer2019automated,gilday2025embodied,ha2021fit2form,kodnongbua2023computational,hiller2012,matthews2023}. 
Recent studies have further applied computational design principles to soft grippers through morphology--control co-design, Bayesian optimization, topology optimization, evolutionary search and so on~\cite{deimel2017automated,Wang2024FinBayesAM,josh2023,jin2024computational}.
In parallel, origami design has likewise benefited from computational methods for generating crease patterns and foldable geometries~\cite{dudte2016programming,demaine2017origamizer,zhao2021computational,Chen2023origami}. S. Kamrava et al.~\cite{sideline} have explicitly proposed a computational design method for origami structures that can be used in gripper design. 
However, few studies have applied computational methods to design customized origami grippers. This gap motivates our learning-based formulation, which optimizes object-specific origami grippers at the design level.


In soft robotics, morphology and actuation are often tightly coupled, and recent bio-inspired systems explicitly exploit this coupling to enhance performance~\cite{huang2026crabedm}.
The generated morphology must also be actuated without sacrificing the simplicity of the structure. 
A range of fabrication and self-folding approaches has been explored for origami structures, including smart materials, magnetic fields, fluidic pressure, and so on~\cite{rus2018design,felton2014method,sideline,li2017fluid,kilian2017string}.
Among them, motorized tendon actuation, which can coordinate distributed folds with few actuators while providing direct and repeatable control, is particularly suitable for origami grippers~\cite{liu2024tendon,Chen2020hopper,Feshbach2023}.
This makes tendon actuation a practical bridge between an optimized crease geometry and a deployable origami gripper.

In this paper, we present an automated computational design framework for designing object-specific, tendon-driven enveloping origami grippers.
Given a target object geometry, our method uses reinforcement learning (RL) in simulation to optimize the morphology of an axisymmetric origami gripper for high-load grasps and converts the resulting design into a manufacturable crease pattern.
We construct the optimized structure through a multi-layer lamination process and a compact motorized tendon-driven mechanism for folding and unfolding.
Experiments on physical grippers across diverse objects validate the proposed design pipeline, demonstrating reliable grasping success and large maximum holding forces.


The main contributions of this work are as follows:
\begin{itemize}
    \item We introduce object-specific computational design into origami grippers, enabling automatic generation of customized enveloping origami grippers from target object geometries.
    \item We establish a physical realization pipeline that spans the fabrication, assembly, and motorized tendon-driven actuation of the origami grippers.
\end{itemize}
\section{Computational Design}
\label{computation}

We parameterize the gripper using axisymmetric assemblies of modified-elbow-pattern-based units~\cite{double-miura}, evaluate candidate morphologies through reduced-order simulation, and formulate object-specific morphology optimization as an RL-based design-refinement process.
\subsection{Geometric Design and Kinematic Modeling}

\begin{figure*}[!t]
    \centering
    \includegraphics[width=0.8\textwidth]{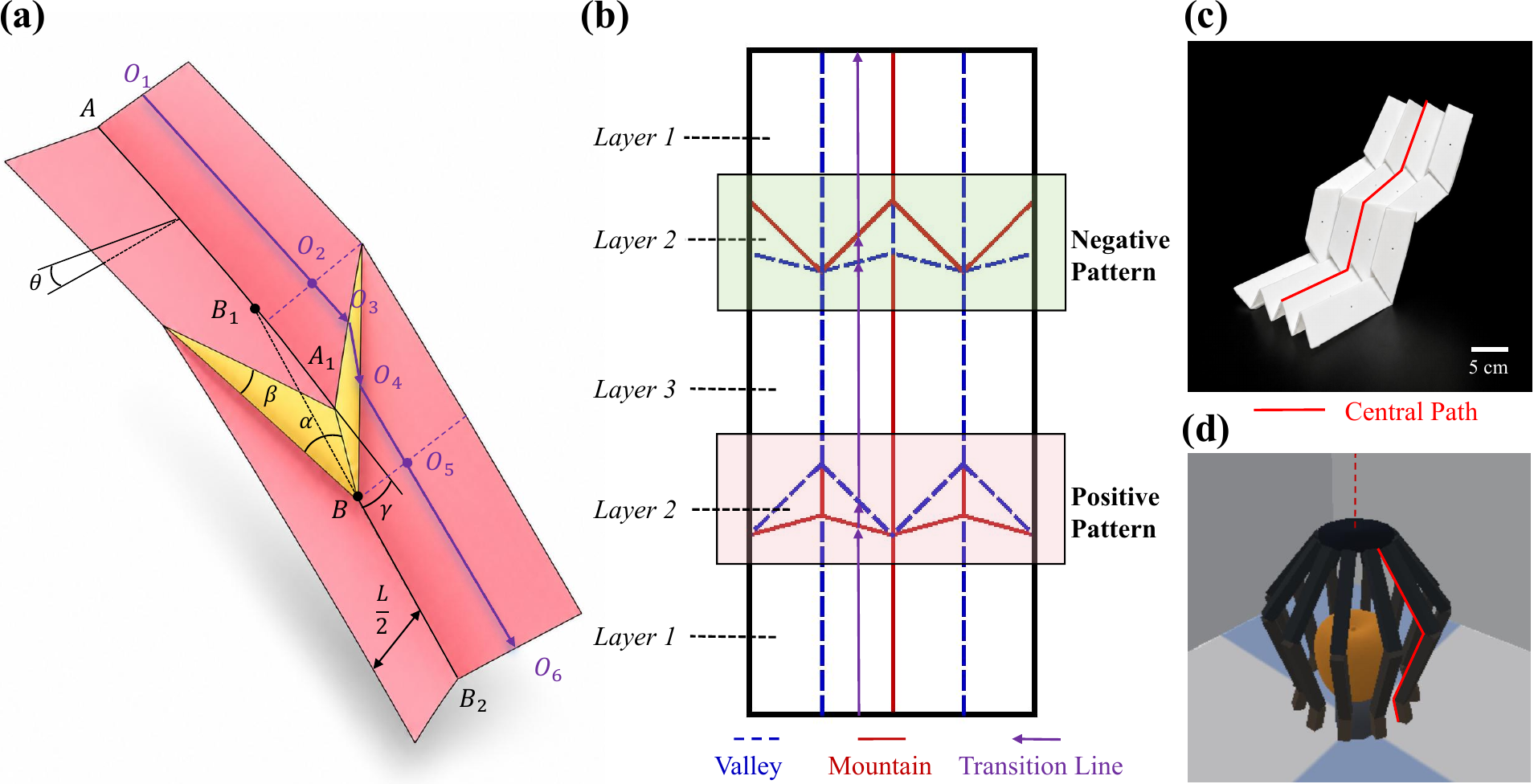}
    \caption{\textbf{Computational design and modeling of the proposed origami gripper.}
    (a) Parametric geometry of the modified elbow-pattern-based unit.
    (b) Crease pattern design with thickness control. 
    (c) Physical prototype with the extracted central path. During grasping, object contact primarily occurs along the folded edges corresponding to the central path.
    (d) Reduced-order linkage model of the axisymmetric gripper in PyBullet, where the twelve fingers are actuated synchronously. This idealization approximates the coordinated closure of the physical gripper (see Secs.~\ref{subsec:fabrication} and~\ref{subsec:tendon_actuation}). 
    }
    \label{fig:geometry}
\end{figure*}

To make the design problem computationally feasible, we optimize the gripper within a selected family of origami units rather than searching over arbitrary crease patterns. 
The four-crease geometry of Miura-ori offers limited local steering for closed-loop enveloping fingers, while Waterbomb shapes remain heavily constrained by the underlying geometry despite richer vertices~\cite{zhao2021computational}.
We therefore adopt modified-elbow-pattern-based units~\cite{double-miura} as the structural primitive type of the gripper. This unit provides sufficient local geometric freedom for enveloping assembly while preserving a broad path-design space for finger deformation.

For each unit, the local geometry is parameterized by the unit length $L$, crease angles $\alpha$ and $\beta$, and folding angle $\theta$, as shown in Fig.~\ref{fig:geometry}(a). The central path of the unit is defined by the polyline $AB_1B_2$, where $AA_1$ and $BB_1$ are the center lines of the two adjacent parts and $B_1$ is their virtual intersection. Under rigid folding, the unit bends the central path by a turning angle $\gamma=\angle A_1B_1B$ computed via:
\begin{equation}
\tan \frac{\gamma}{2}
=
\sin\theta \tan\beta .
\label{eq:unit_turning_angle}
\end{equation}
Thus, for a prescribed design folding angle $\theta^\ast$, a desired local path turn $\gamma$ determines the unit angle as
\begin{equation}
\beta
=
\arctan\left(
\frac{\tan(\gamma/2)}{\sin\theta^\ast}
\right).
\label{eq:inverse_beta}
\end{equation}

To rapidly obtain the corresponding crease pattern, we convert the central path to the transition-line representation, which can be automatically transferred to the corresponding crease pattern~\cite{yang2024design}. As shown in Fig.~\ref{fig:geometry}(a), the lengths of $O_1O_2$ and $O_5O_6$ are respectively equal to the lengths of the linkage segments before and after the unit, while the middle portion $O_2O_3O_4O_5$ is determined by the local unit geometry:
\begin{equation}
\begin{aligned}
\left|O_2O_3\right|
&=
\frac{L}{4\tan(\alpha+\beta)}, \\
\left|O_3O_4\right|
&=
\frac{L}{4}
\left(
\frac{1}{\tan\alpha}
-
\frac{1}{\tan(\alpha+\beta)}
\right), \\
\left|O_4O_5\right|
&=
\frac{L}{4\tan\alpha}.
\end{aligned}
\label{eq:transition_line_unit}
\end{equation}
Equations~\eqref{eq:inverse_beta}--\eqref{eq:transition_line_unit} define the local conversion from a prescribed central-path turn to the transition-line segment.

We further address finite panel thickness in unit assembly. As shown in Fig.~\ref{fig:geometry}(b), a unit, denoted as the positive pattern, introduces a three-layer offset along the thickness direction. Its mirrored counterpart, denoted as the negative pattern, reverses this offset and compensates for layer accumulation. We therefore alternate positive and negative units along each finger, which limits the maximum number of overlapping panel layers to three to preserve practical foldability.

These local relations define a direct path-to-pattern workflow. The target finger is first represented as a serial linkage, whose polyline specifies the desired central path of the folded origami finger. For each finger, the unit orientation is selected according to the thickness rule above, and Eq.~\eqref{eq:inverse_beta} determines the unit parameter.
Then Eq.~\eqref{eq:transition_line_unit} generates the corresponding transition-line segment. Concatenating these local segments yields the full transition line of the finger, which is then passed to the origami design software presented in~\cite{yang2024design} to generate the complete planar crease layout.
\subsection{Simulation Model for Gripper Design}
To enable efficient design optimization, we build a reduced-order simulation model of the proposed enveloping origami gripper in PyBullet~\cite{pybullet} as shown in Fig.~\ref{fig:geometry}(d).

Rather than simulate the full crease-panel mechanics, we model each origami finger as a planar articulated linkage derived from the central-path polyline. Each polyline segment is represented as a rigid link, and adjacent segments are connected by revolute joints. This abstraction preserves the dominant bending geometry of the origami finger while avoiding the high computational cost of modeling individual panels, creases, and folding dynamics.
The axisymmetric morphology of the gripper is modeled by uniformly arranging $N=12$ identical fingers around a circular palm.

In the simulation, each gripper is initialized in a $60\%$ pre-folded configuration. To reduce the dimensionality of the design space, we fix $\theta=35^\circ$, $\alpha=45^\circ$, and $L=51.75~\mathrm{mm}$ as predefined geometric hyperparameters throughout optimization. The remaining morphology parameters are optimized in the subsequent learning-based design stage.
\subsection{Learning-Based Design Optimization}
For a fixed target object, we formulate the morphology search of the proposed gripper as a finite-horizon Markov decision process (MDP). The policy learned here acts as a design optimizer: it observes the current gripper morphology and the previous design update in simulation and computes updates to the gripper design parameters. The target object is fixed in the simulation environment and influences the optimization through grasping outcomes and the corresponding rewards. The framework centers on simulation-based morphology optimization rather than a particular search algorithm.
Here, we use Proximal Policy Optimization (PPO)~\cite{ppo} to learn sequential design refinements from simulation-based rewards. The key components of the MDP formulation are defined as follows.
\paragraph{Observation Space}
The observation encodes the current polyline state and the previous design update. We first define the polyline state as:
\begin{equation}
\mathbf{P}_t =
\left[
(x_1, y_1, e_1),
\ldots,
(x_n, y_n, e_n)
\right]
\in \mathbb{R}^{3n},
\end{equation}
where $n$ is the maximum number of vertices, specified in advance and set to $10$ by default.
Each candidate polyline vertex is represented by $(x_i, y_i, e_i)$, where $(x_i,y_i)$ denotes its planar position and $e_i \in [-1,1]$ is an existence flag indicating whether the point is active. 
A point is treated as active only when $e_i>0$; inactive points are ignored when constructing the central-path polyline. 
To provide temporal information about the previous design step, the previous action $\mathbf{a}_{t-1}$ is appended to the observation, where $\mathbf{a}_{0}=\mathbf{0}$. 
The actor and critic receive the same observation:
\begin{equation}
\mathbf{o}_t =
\left[
\mathbf{P}_t,
\mathbf{a}_{t-1}
\right]
\in \mathbb{R}^{6n}, 
\end{equation}
This fixed-dimensional representation allows PPO to optimize a variable-topology polyline by changing both the positions and activation states of candidate control points.
\paragraph{Action Space}
The action controls the geometric update of the polyline. 
For each control point, the policy outputs normalized actions
$(a^x_{i,t},a^y_{i,t},a^e_{i,t})\in[-1,1]^3$.
The state update is
\begin{equation}
\begin{aligned}
x_{i,t+1}
&=\operatorname{clip}(x_{i,t}+0.03a^x_{i,t},-0.08,0.08),\\
y_{i,t+1}
&=\operatorname{clip}(y_{i,t}+0.03a^y_{i,t},0,0.15),\\
e_{i,t+1}
&=a^e_{i,t}.
\end{aligned}
\end{equation}

\paragraph{Reward Function}
The reward evaluates each candidate morphology in terms of grasping performance and physical feasibility:
\begin{equation}
    r_t =
    r_{\mathrm{grasp}}
    +
    p_{\mathrm{feas}}
    +
    r_{\mathrm{bonus}}.
\end{equation}

The grasping term combines sparse task success with dense lifting progress, while the feasibility penalty discourages geometrically or physically infeasible morphologies:
\begin{equation}
    r_{\mathrm{grasp}}
    =
    w_{\mathrm{succ}} r_{\mathrm{succ}}
    +
    w_{\mathrm{height}} r_{\mathrm{height}},
\end{equation}
\begin{equation}
    p_{\mathrm{feas}}
    =
    w_{\mathrm{len}} p_{\mathrm{len}}
    +
    w_{\mathrm{ang}} p_{\mathrm{ang}}
    +
    w_{\mathrm{gnd}} p_{\mathrm{gnd}},
\end{equation}

We weight grasping success more heavily than partial lifting progress, while retaining the height reward to guide intermediate improvements.
The length and angle penalties discourage geometries that exceed the available PVC sheet dimensions or lead to difficult-to-fabricate folds.
Together with the ground-contact penalty and feasibility bonus, these terms guide the search toward successful, manufacturable designs. The details are shown in Table~\ref{tab:reward}.

\begin{table}[t]
\centering
\caption{Objective terms for design optimization.}
\label{tab:reward}
\scriptsize
\setlength{\tabcolsep}{3.5pt}
\renewcommand{\arraystretch}{1.08}
\begin{tabularx}{\columnwidth}{@{}l c X@{}}
\toprule
Term & Weight & Definition \\
\midrule
$r_{\mathrm{succ}}$ 
& $+20$ 
& $1$ if the object is lifted above $85\% h_{\mathrm{tar}}$; $0$ otherwise. \\

$r_{\mathrm{height}}$ 
& $+8$ 
& $\mathrm{clip}\!\left(h_{\mathrm{lift}}/h_{\mathrm{tar}},\,0,\,1\right)$, with $h_{\mathrm{tar}}=0.2\,\mathrm{m}$. \\

$p_{\mathrm{len}}$ 
& $-5$ 
& $\max\!\left(0,\,\sum_i l_i - l_{\max}\right)\times 100$, with $l_{\max}=0.32\,\mathrm{m}$. \\

$p_{\mathrm{ang}}$ 
& $-5$ 
& Number of local turns with $|\theta_i|<20^\circ$ or $|\theta_i|>75^\circ$. \\

$p_{\mathrm{gnd}}$ 
& $-10$ 
& $1$ if undesired ground contact occurs during closure; $0$ otherwise. \\

$r_{\mathrm{bonus}}$ 
& \textemdash 
& $4$ if $r_{\mathrm{succ}}=1$ and $p_{\mathrm{len}}=p_{\mathrm{ang}}=p_{\mathrm{gnd}}=0$; $0$ otherwise. \\
\bottomrule
\end{tabularx}
\end{table}

\subsection{Training Protocol}

Training proceeds as iterative morphology refinement in simulation. At each decision step, the policy updates the current central-path polyline. The simulator then resets the scene, constructs the corresponding reduced-order gripper, loads the target object mesh, and initializes the gripper at a fixed height $h$ above the object center of mass. 
All fingers close synchronously around the object, after which the gripper is lifted upward to evaluate the candidate morphology.

At the beginning of each design-refinement step, the gripper and target object are reset to the same predefined initial poses.
Each episode contains a sequence of such steps. It terminates when the maximum number $S=120$ of refinement steps is reached or when the policy produces a design that successfully grasps the object without triggering any feasibility penalty.

We employ a mass-based curriculum to progressively increase the difficulty of morphology optimization. At curriculum stage $k$, the object mass is fixed at $m_k$. Each step generates one candidate gripper and performs one grasping trial, and $\bar{s}_k$ is defined as the average success rate over the most recent $10{,}000$ trials at the current mass level. When $\bar{s}_k$ exceeds $\eta=0.35$, the object mass is increased:

\begin{equation}
m_{k+1} =
\begin{cases}
\min(m_k+\Delta m, m_{\max}), & \bar{s}_k > \eta,\\
m_k, & \bar{s}_k \leq \eta.
\end{cases}
\label{eq:mass_curriculum}
\end{equation}
By default, $m_0=1$~kg, $\Delta m=1$~kg, and $m_{\max}=25$~kg.
Object geometry remains fixed throughout the curriculum.
The upper bound is an empirical training ceiling rather than a physical payload rating.

 


\section{Fabrication}
\label{fab}

\begin{figure*}[!t]
    \centering
    \includegraphics[width=0.85\textwidth]{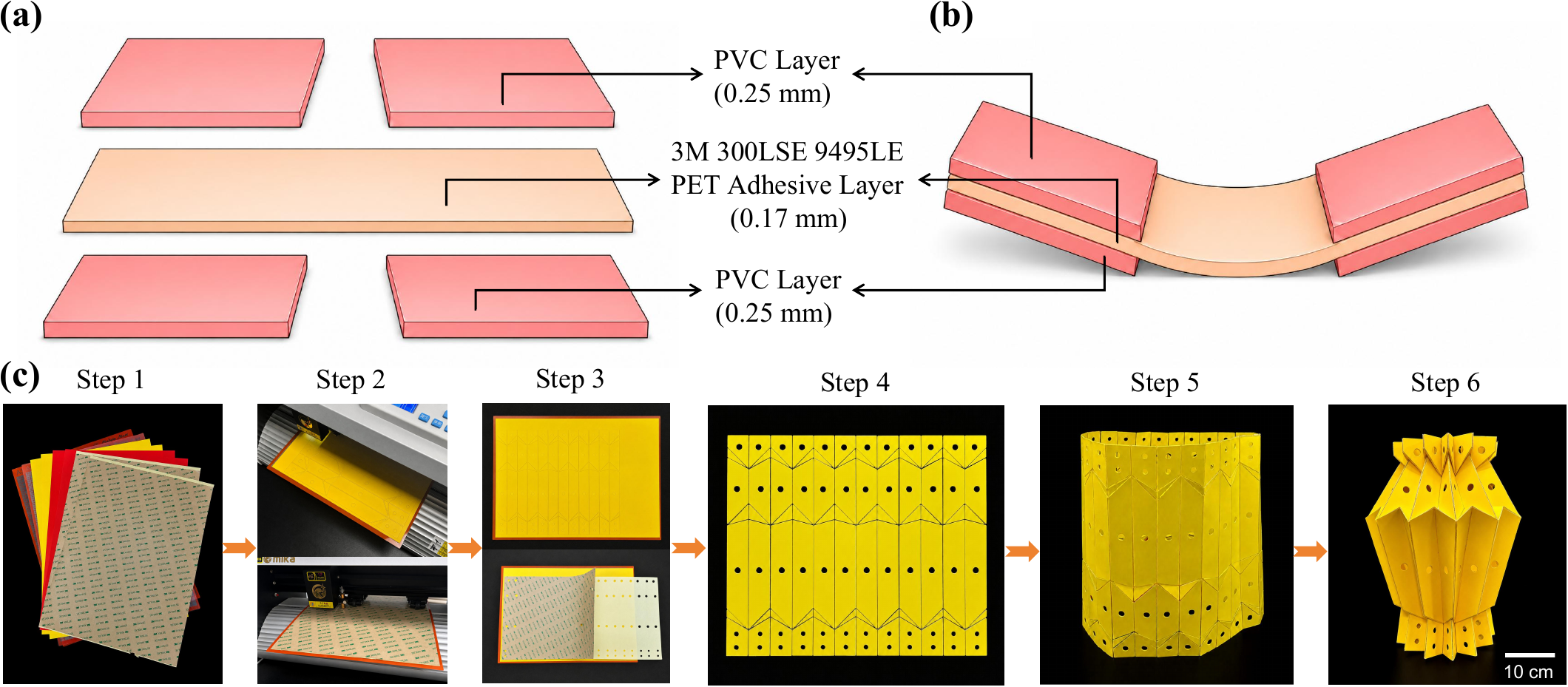}
    \caption{
\textbf{Fabrication process of the customized origami gripper.}
(a) Laminated sandwich structure.
(b) Schematic diagram of the folded structure after assembly.
(c) Fabrication workflow. Step 1: material preparation; Steps 2--4: cutting and laminating the PVC and adhesive layers into a composite origami sheet; Step 5: connect two fabricated sheets end-to-end to form a cylindrical gripper precursor; Step 6: manually pre-fold the structure.
}
    \label{fig:fabrication}
\end{figure*}

\subsection{Structural Design and Material Selection}
\label{subsec:sandwich_model}

To fabricate the optimized crease patterns, we adopt a laminated sandwich structure, a widely used strategy in origami-inspired robots and grippers, where layered sheets combine stiff panels with foldable hinge regions~\cite{felton2014method,magicball}. 
In our design, this structure is simplified into three layers: two outer structural layers and one compliant intermediate layer as shown in Fig.~\ref{fig:fabrication}(a) and (b). 
The outer layers provide the strength required for load-bearing origami panels, while the intermediate layer maintains bonding between the panels and accommodates folding at the crease regions.

In the prototype, the intermediate layer is a double-sided adhesive tape that bonds the outer sheets while allowing localized bending. For the outer structural layers, we compared 3D-printed PLA panels with cut PVC sheets. PLA was initially considered because it can directly realize custom panel geometries, but printing a complete gripper takes 20 h. In contrast, PVC sheets are fabricated by simple planar cutting and lamination in 2 h and show better toughness in practice, making them more suitable for rapid, standardized fabrication. Therefore, PVC is selected for the outer layers.

\subsection{Manufacturing and Assembly Workflow}
\label{subsec:fabrication}

Fig.~\ref{fig:fabrication}(c) summarizes the fabrication and assembly workflow that converts the optimized planar crease pattern into a tendon-driven origami gripper. 
The gripper body is fabricated as a laminated thick-panel sheet consisting of two PVC structural layers and an intermediate PET layer with double-sided adhesive. 
The PVC layers provide panel strength, while the adhesive PET bonds the laminate and serves as the compliant interlayer required for repeated folding.

The optimized pattern is exported as digital cutting paths. 
For each PVC layer, the prescribed crease network is cut to define the folding lines, and tendon-routing holes are cut at selected cell centroids. 
For the adhesive sheet, only the corresponding routing holes are cut, leaving the adhesive region otherwise continuous to maintain reliable interlayer bonding. 
The three layers are laminated according to the thick-panel model introduced in Sec.~\ref{subsec:sandwich_model}. 
After lamination, the composite sheet is kept under static pressure to improve bonding uniformity and prevent delamination during folding.

The bonded planar sheets are subsequently assembled into the gripper body. 
Two laminated sheets are wrapped and joined along their lateral boundaries to form a closed cylindrical structure consistent with the axisymmetric design. 
Finally, the cylinder is manually pre-folded along the prescribed mountain and valley creases, forming a 12-finger gripper tessellated from elbow-pattern-based units, consistent with the $N=12$ morphology in simulation. The driving tendons are threaded through the centroid holes following the tendon-routing path. The material cost per gripper, excluding the motor, is less than US\$5.
This standardized workflow converts optimized crease patterns into physical grippers compatible with motorized tendon actuation.
\subsection{Tendon Routing and Motorized Actuation}
\label{subsec:tendon_actuation}

\begin{figure}[!t]
    \centering
    \includegraphics[width=0.9\columnwidth]{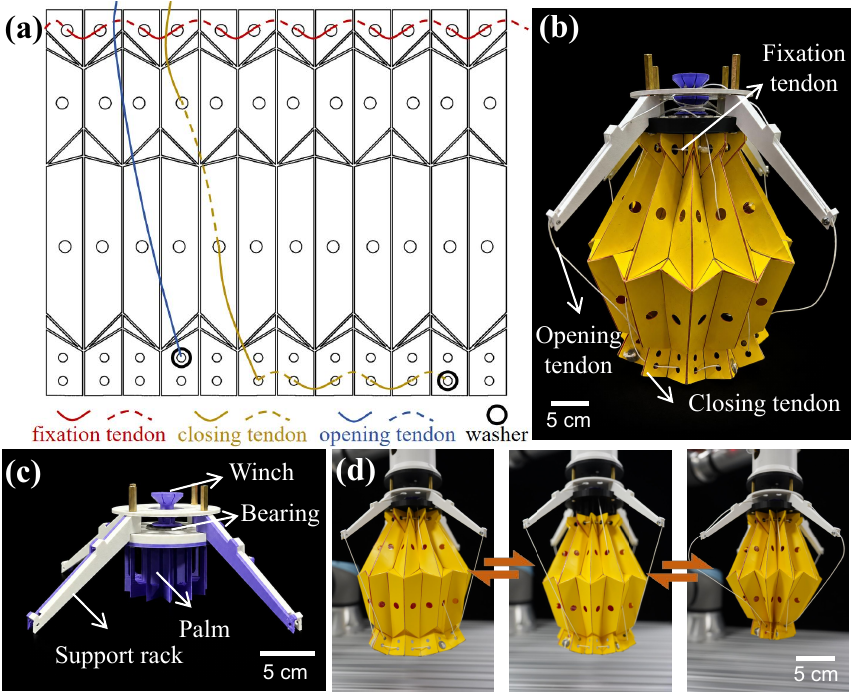}
    \caption{\textbf{Tendon routing and actuation system of the origami gripper.}
    (a) Schematic of the tendon-routing layout. Solid and dashed lines indicate tendon segments routed above and below the origami surface, respectively.
    (b) Physical implementation of the tendon-routing scheme.
    (c) Palm and actuation mechanism.
    (d) A single motor (GB37-3625 brushless DC motor, 24\,V, 7\,rpm, rated output torque: 3.43\,\text{N$\cdot$m}) bidirectionally actuates the gripper, closing it clockwise and opening it counterclockwise.}
    \label{fig:string}
\end{figure}

Fig.~\ref{fig:string} shows the tendon-routing scheme and the corresponding actuation mechanism. 
The gripper uses three groups of tendons: a tendon for fixation, four closing tendons, and four opening tendons. All tendons are braided, high-strength polyethylene (PE) tendons with a diameter of 1 mm.

The fixation tendon is routed around the top perimeter of the origami cylinder to secure the gripper to the palm, while the other two groups form an antagonistic tendon transmission for active folding (closing) and unfolding (opening).

The two active tendon groups follow complementary routing paths. 
Each closing tendon is connected to the lower layer of a coaxial winch (diameter: 16\,mm), routed through prescribed holes across neighboring cylinder generatrices, and anchored near the distal ring after spanning approximately one quarter of the circumference. 
Each opening tendon is connected to the upper layer of the winch, guided outward by a support rack, and anchored to the distal origami panels through the bottom holes. 
For both active tendon groups, the distal end of each tendon passes through a washer and is tied with a knot, so that the tensile load is transferred to the terminal cell over a finite contact area. 
When tensioned, the closing tendons pull the distal origami units inward and drive the cylindrical body toward its folded grasping configuration. 
Conversely, tensioning the opening tendons pulls the terminal panels outward and induces unfolding for release.

The two active tendon groups are wound in opposite directions on the two layers of the same winch. 
Thus, one motor direction tensions the closing tendons and releases the opening tendons, whereas reverse rotation interchanges their roles. 
This double-layer antagonistic transmission enables both grasping and release with a single motor, while keeping the palm compact and simple.
\section{Results}
\subsection{Design Results}
\label{subsec:results}
 
We apply the pipeline to four YCB objects~\cite{ycb}, including three geometrically distinct fruits and a flattened, elongated bottle: apple, peach, pear, and mustard bottle. 

\begin{figure}[t]
    \centering
    \includegraphics[width=0.9\columnwidth]{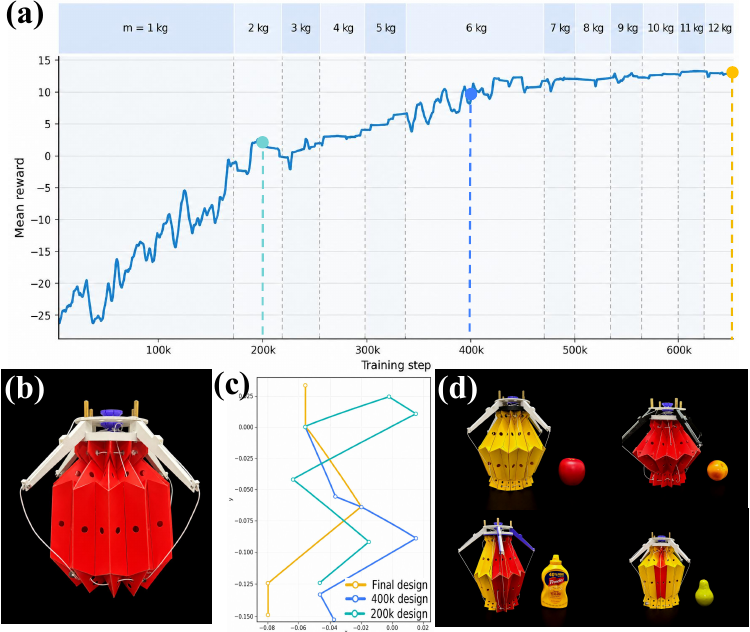}
    \caption{\textbf{Computational design results.}
    (a) Training process of the apple gripper.
    (b) Unoptimized baseline.
    (c) Evolution of finger shape with maximum liftable masses of 2.5, 8, and 12 kg in separate simulation tests.
    (d) Optimized grippers for apple, peach, mustard bottle, and pear.}
    \label{fig:results}
\end{figure}

Fig.~\ref{fig:results}(a) shows the training process for the apple gripper. Fig.~\ref{fig:results}(c) illustrates the evolution of gripper morphology through representative designs generated by the policy at different training stages, showing a progression toward simpler, more compact, and more effective enveloping shapes. The optimized designs for all four objects are then converted into crease patterns and fabricated following Sec.~\ref{fab}.



\subsection{Experimental Setup and Evaluation Method}
We evaluated the proposed computational design and fabrication framework through three complementary questions:
\begin{enumerate}
    \item Can an object-specific origami gripper reliably grasp its target in the real world?
    \item What maximum holding force can the customized gripper achieve on its target object?
    \item How does a customized gripper perform on objects other than its design target?
\end{enumerate}
So we conducted target object grasping experiments, maximum holding-force evaluations, and cross-object force tests.


The target object grasping experiments were performed with a UR5e arm and the corresponding physical YCB objects, testing whether each customized gripper could reliably grasp its intended target in the real world.
The force-characterization experiments were conducted on an Instron 68TM-10 universal testing machine. 
For these tests, the objects were 3D printed from the official YCB STL models, with an additional mounting bar for fixation to the Instron clamp. 
The same setup was used for both maximum holding force evaluation on the target object and cross-object validation on non-target objects, allowing us to characterize both grasping capacity and object specificity.

\subsection{Target Object Grasping Experiments}
\label{grasp_exp}
\begin{figure}[!t]
    \centering
    \includegraphics[width=0.95\columnwidth]{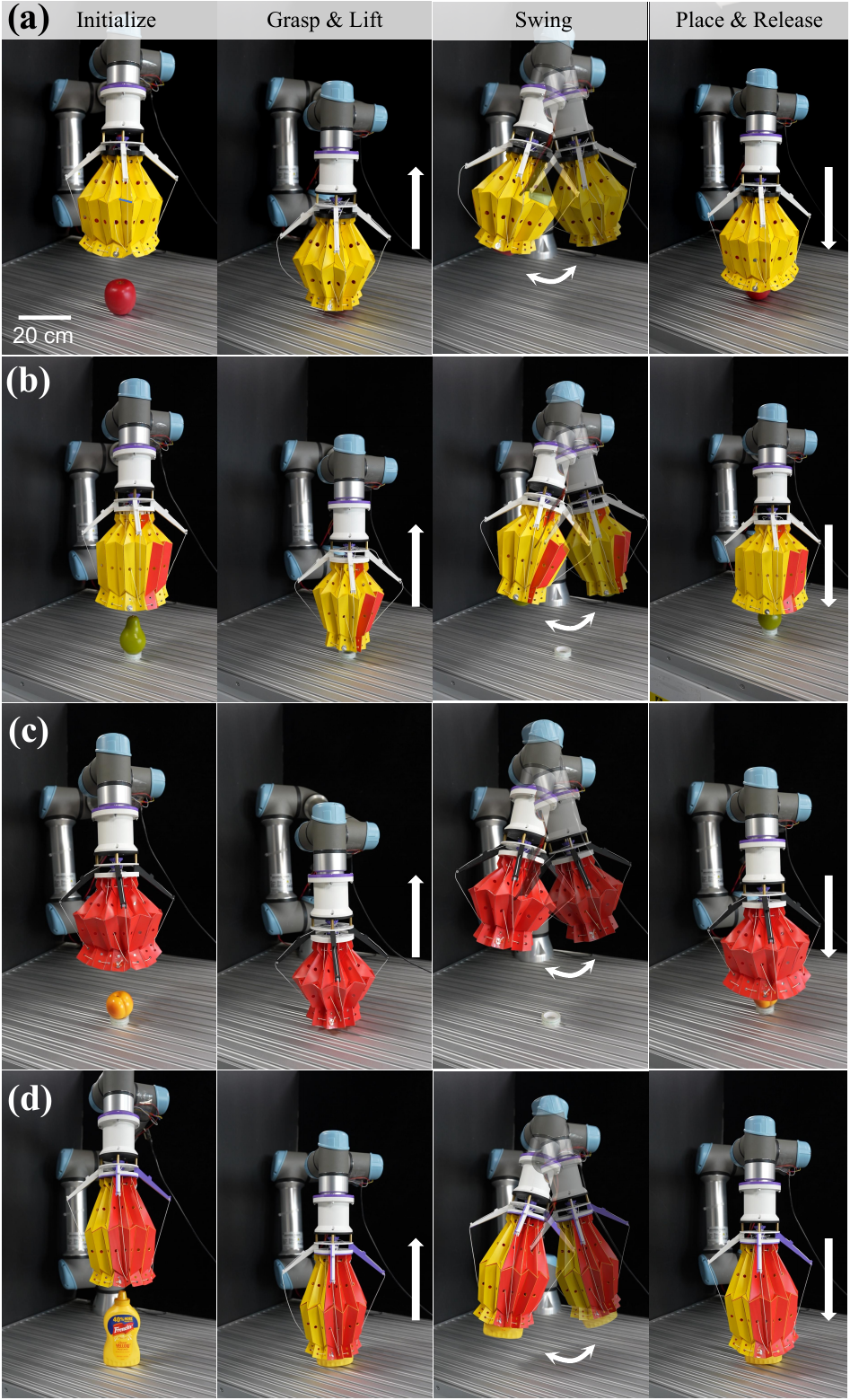}
    \caption{\textbf{YCB object grasping experiments.}
    (a) Apple (64.83\,g); (b) Pear (53.21\,g); (c) Peach (34.01\,g); (d) Mustard bottle (45.57\,g).}
    \label{fig:grasp_demo}
\end{figure}

We first evaluated whether each customized gripper could reliably grasp its intended target in a robotic manipulation setting. 
Each gripper was mounted on a UR5e arm and tested on the corresponding physical YCB object. 
As shown in Fig.~\ref{fig:grasp_demo}, in each trial, the robot lowered the gripper to a predefined grasping pose, actuated the closing tendons to secure the object, lifted the object by approximately $20~\mathrm{cm}$, and then performed a lateral swing of approximately $\pm 30^{\circ}$ to test grasp stability. 
A trial was deemed successful if the object remained secured throughout lifting and swinging and was safely placed back without dropping.

We conducted 10 trials per target object under nominal grasping conditions, with all 40 trials successful.
The objects remained secured during lifting and swinging, demonstrating grasp stability under the tested conditions.
\begin{figure}[!t]
    \centering
    \includegraphics[width=\columnwidth]{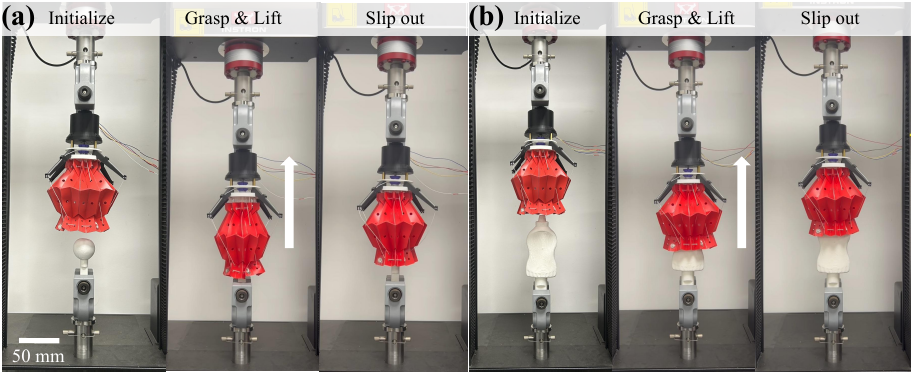}
    \caption{\textbf{Maximum holding force evaluation.}
    (a) The peach gripper grasps the peach (target-object grasping).
    (b) The peach gripper grasps the mustard bottle (cross-object grasping).}
    \label{fig:instron_demo}
\end{figure}


\subsection{Maximum Holding Force Evaluation}
\label{max_exp}
As shown in Fig.~\ref{fig:instron_demo}, we further quantified the force capacity of each customized gripper using an Instron 68TM-10 universal testing machine. 
The gripper was fully closed around the printed object, after which the Instron crosshead moved upward at a constant speed of $2~\mathrm{mm/s}$ while the axial load was recorded, yielding a quasi-static pull-out force curve.
The test was stopped after the force dropped sharply, indicating detachment between the gripper and the object.

\begin{figure}[!t]
    \centering
    \includegraphics[width=\columnwidth]{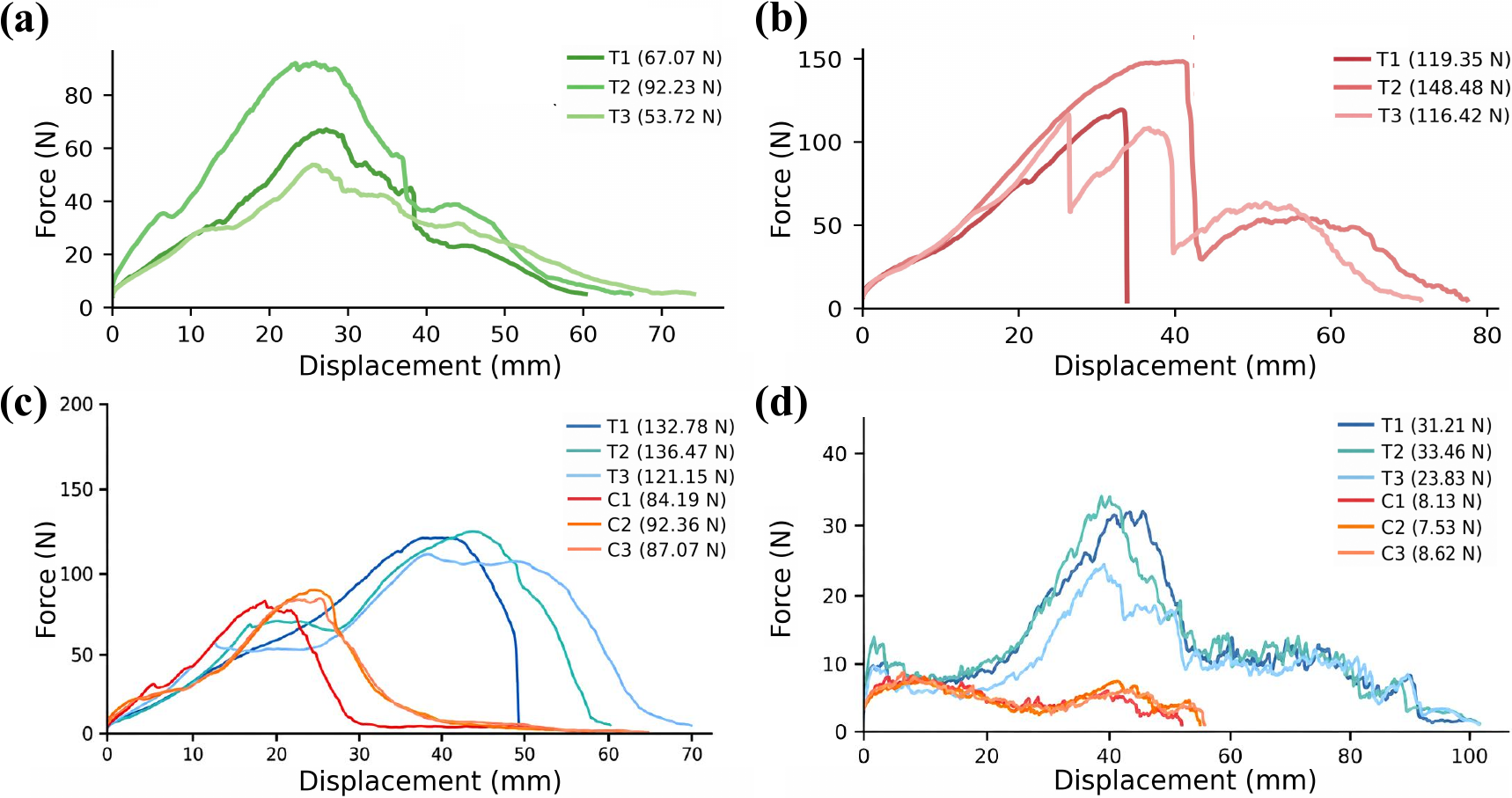}
    \caption{
\textbf{Force-displacement curves of customized origami grippers.} Each experiment was repeated three times.
(a) Pear-specific gripper grasps the pear.
(b) Apple-specific gripper grasps the apple.
(c) Comparison between the peach-specific gripper (T1-T3) and the bottle-specific gripper (C1-C3) when grasping a peach.
(d) Comparison between the bottle-specific gripper (T1-T3) and the peach-specific gripper (C1-C3) when grasping a bottle.
}
    \label{fig:exp_force}
\end{figure}

For each trial, the maximum holding force $F_{\max}$ is defined as the peak value of the quasi-static pull-out force curve before detachment.
Each customized gripper was tested three times on its target object.
To evaluate the benefit of morphology optimization, we also tested an unoptimized baseline gripper corresponding to the initial design used in PPO optimization, as shown in Fig.~\ref{fig:results}(b).
Representative force curves are shown in Fig.~\ref{fig:exp_force}, and the averaged maximum holding forces are summarized in Table~\ref{tab:max_holding_force}.


The tests demonstrate that the customized grippers sustain large holding forces on their intended targets while maintaining high force-to-weight ratios.
This advantage is most pronounced for rounded fruit-like objects, consistent with the gripper's intrinsic axisymmetric enveloping morphology. While performance is comparatively lower on the flatter bottle geometry, the gripper still achieves substantial holding force, indicating that the morphology retains useful adaptability beyond its most favorable object shape.

The baseline comparison demonstrates that object-specific morphology optimization enhances holding capacity under otherwise unchanged material, fabrication, and actuation conditions.
Together, these results show how geometry-programmed folding and target-specific enclosure can turn thin laminated sheets into strong grippers while retaining a simple actuation mechanism.
\subsection{Cross-Object Grasping Validation}
\label{cross_exp}

\begin{table}[t]
\centering
\caption{Gripper performance: baseline versus optimized designs.}
\label{tab:max_holding_force}
\begingroup
\footnotesize
\setlength{\tabcolsep}{3pt}
\renewcommand{\arraystretch}{1.15}
\newcommand{\paperforce}[2]{\ensuremath{#1}\,{\scriptsize\color{black!55}\ensuremath{\pm #2}}}
\begin{tabular}{@{}lccc@{\hspace{6pt}}ccc@{}}
  \toprule
  \textsf{\textbf{Object}} &
  \multicolumn{3}{c}{\textsf{\textbf{Baseline}}} &
  \multicolumn{3}{c@{}}{\textsf{\textbf{Optimized designs}}} \\
  \cmidrule(lr){2-4}\cmidrule(l){5-7}
  & \makecell{$m_g$\\{[g]}} & \makecell{$F_{\max}$\\{[N]}} & $\dfrac{F_{\max}}{m_g g}$
  & \makecell{$m_g$\\{[g]}} & \makecell{$F_{\max}$\\{[N]}} & $\dfrac{F_{\max}}{m_g g}$ \\
  \midrule
  Pear   & 168.84 & \paperforce{5.01}{0.74} & 3.0 & 173.20 & \paperforce{\mathbf{71.01}}{15.96} & $\mathbf{41.0}$ \\
  Apple  & 168.84 & \paperforce{9.46}{1.12} & 5.6 & 160.76 & \paperforce{\mathbf{128.08}}{14.47} & $\mathbf{79.7}$ \\
  Peach  & 168.84 & \paperforce{6.80}{1.70} & 4.0 & 156.40 & \paperforce{\mathbf{130.13}}{6.53} & $\mathbf{83.2}$ \\
  Bottle & 168.84 & \paperforce{5.77}{0.74} & 3.4 & 176.34 & \paperforce{\mathbf{29.50}}{4.11} & $\mathbf{16.7}$ \\
  \bottomrule
\end{tabular}\par
\smallskip
{\scriptsize\color{black!55}$m_g$ excludes the motor and arm adapter (407.1\,g combined); $g=10\,\mathrm{m\,s^{-2}}$.\par}
\endgroup
\end{table}

\begin{figure}[!t]
    \centering
    \includegraphics[width=\columnwidth]{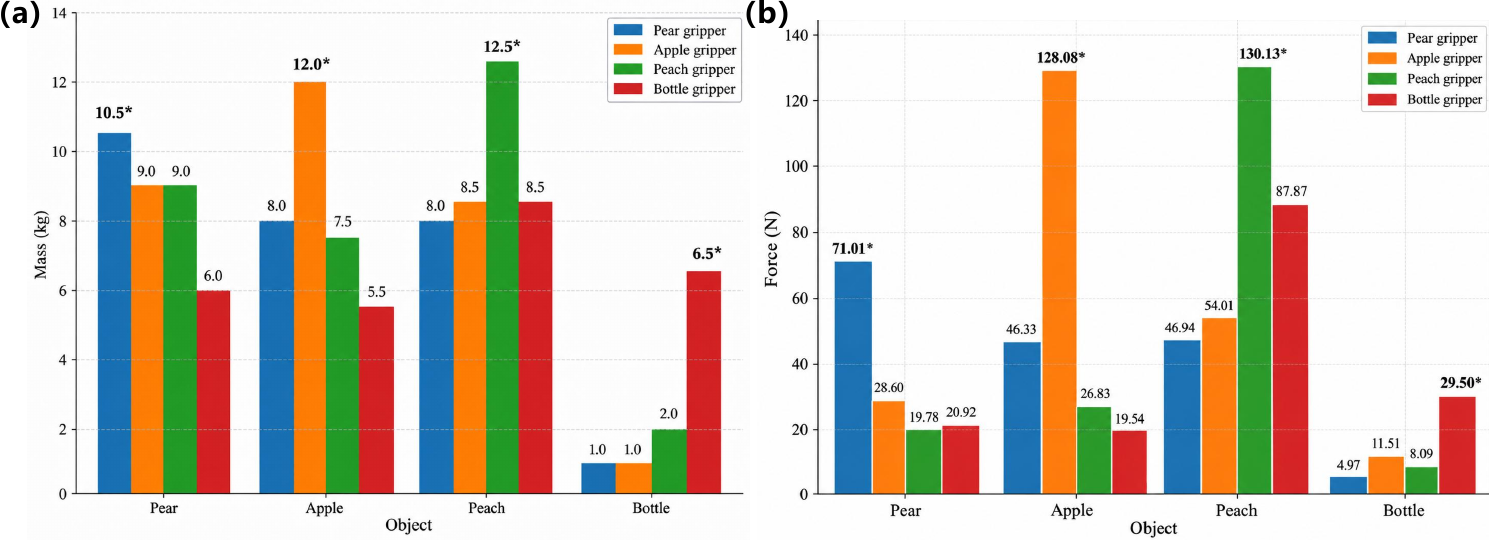}
    \caption{Cross-object grasping evaluation. (a) Simulated maximum lifted mass (kg). (b) Measured maximum holding force (N).}
    \label{fig:cross_object_matrix}
\end{figure}


Finally, we evaluated whether the measured force capacity arose from object-specific morphological matching rather than generic compliance alone.
We performed cross-object tests among all four customized grippers and all four target objects in both simulation and physical experiments.
In simulation, we progressively increased the object mass in 0.5 kg increments until the gripper failed to lift it, recording the maximum successfully lifted mass.
In physical experiments, the same Instron protocol described in Sec.~\ref{max_exp} was used to measure the maximum holding force, as shown in Fig.~\ref{fig:instron_demo}.

The cross-object results are shown in Fig.~\ref{fig:cross_object_matrix}.
Both simulation and physical experiments exhibit a consistent target-specific advantage: for each object, its customized gripper achieves the highest simulated liftable mass and measured mean holding force, respectively.
Despite the simplifications of the reduced-order simulation and the quantitative discrepancies with physical experiments, both exhibit strong agreement in identifying the best-performing gripper morphology for each target object.

These results indicate that the optimized crease morphology encodes object-specific geometric compatibility. 
Rather than acting as a purely universal compliant gripper, the proposed design leverages object-specific shape matching to improve holding force on the intended target.
\section{Conclusion}
We presented a complete framework for designing and fabricating customized tendon-driven enveloping origami grippers from object geometry. The framework combines learning-based morphology optimization, laminated-sheet fabrication, and compact single-motor tendon actuation, thereby translating an object mesh into a manufacturable customized gripper morphology.

Physical experiments across multiple YCB objects validate that the resulting grippers can reliably grasp their intended targets and generate large, stable holding forces. The cross-object evaluation further shows that grasp performance depends critically on the geometric compatibility between the gripper morphology and the object. In particular, the optimized axisymmetric structures leverage distributed enclosure and contact to convert lightweight sheet materials into high-capacity end-effectors.

The fixed axisymmetric architecture may limit performance on strongly non-axisymmetric objects, while grasping robustness across a broad range of object poses remains to be established.
Manufacturing consistency across independent builds and long-term durability have not been systematically evaluated.
Future work will explore more general origami morphologies, optimization over broader object-pose distributions, and systematic repeatability and durability testing.


\bibliographystyle{IEEEtran}
\bibliography{references}

@inproceedings{long2020systematic,
  title={A systematic review and meta-analysis of robotic gripper},
  author={Long, Zhang and Jiang, Qian and Shuai, Tao and Wen, Feijuan and Liang, Chunping},
  booktitle={IOP Conference Series: Materials Science and Engineering},
  volume={782},
  number={4},
  pages={042055},
  year={2020},
  organization={IOP Publishing}
}

@article{gilday2025embodied,
  title={Embodied manipulation with past and future morphologies through an open parametric hand design},
  author={Gilday, Kieran and Sirithunge, Chapa and Iida, Fumiya and Hughes, Josie},
  journal={Science Robotics},
  volume={10},
  number={102},
  pages={eads6437},
  year={2025},
  publisher={American Association for the Advancement of Science}
}

@article{rus2015design,
  title={Design, fabrication and control of soft robots},
  author={Rus, Daniela and Tolley, Michael T},
  journal={Nature},
  volume={521},
  number={7553},
  pages={467--475},
  year={2015},
  publisher={Nature Publishing Group UK London}
}

@article{rus2018design,
  title={Design, fabrication and control of origami robots},
  author={Rus, Daniela and Tolley, Michael T},
  journal={Nature Reviews Materials},
  volume={3},
  number={6},
  pages={101--112},
  year={2018},
  publisher={Nature Publishing Group UK London}
}

@article{mahadevan2005self,
  title={Self-organized origami},
  author={Mahadevan, Lakshminarayanan and Rica, Sergio},
  journal={Science},
  volume={307},
  number={5716},
  pages={1740--1740},
  year={2005},
  publisher={American Association for the Advancement of Science}
}

@ARTICLE{jin2024computational,
  author={Jin, Jiayi and Feng, Siyuan and Li, Shuguang},
  journal={IEEE Robotics and Automation Letters}, 
  title={Computational Design of Customized Vacuum-Driven Soft Grippers}, 
  year={2025},
  volume={10},
  number={2},
  pages={1641-1648},
  doi={10.1109/LRA.2024.3523203}}

@article{liu2024tendon,
  title={Tendon driven bistable origami flexible gripper for high-speed adaptive grasping},
  author={Liu, Weirui and Bai, Xu and Yang, Hui and Bao, Ruixin and Liu, Jiaqi},
  journal={IEEE Robotics and Automation Letters},
  volume={9},
  number={6},
  pages={5417--5424},
  year={2024},
  publisher={IEEE}
}

@inproceedings{schroeffer2019automated,
  title={An automated design approach for task-specific two finger grippers for industrial applications},
  author={Schroeffer, Andreas and Rehekampff, Christoph and Lueth, Tim C},
  booktitle={2019 IEEE International Conference on Robotics and Biomimetics (ROBIO)},
  pages={184--189},
  year={2019},
  organization={IEEE}
}

@article{kodnongbua2023computational,
author = {Kodnongbua, Milin and Good, Ian and Lou, Yu and Lipton, Jeffrey and Schulz, Adriana},
title = {Computational design of passive grippers},
year = {2022},
issue_date = {July 2022},
publisher = {Association for Computing Machinery},
address = {New York, NY, USA},
volume = {41},
number = {4},
issn = {0730-0301},
url = {https://doi.org/10.1145/3528223.3530162},
doi = {10.1145/3528223.3530162},
journal = {ACM Trans. Graph.},
month = jul,
articleno = {149},
numpages = {11}
}

@inproceedings{ha2021fit2form,
  title={Fit2Form: 3D generative model for robot gripper form design},
  author={Ha, Huy and Agrawal, Shubham and Song, Shuran},
  booktitle={Conference on Robot Learning},
  pages={176--187},
  year={2021},
  organization={PMLR}
}

@inproceedings{demaine2017origamizer,
author = {Demaine, Erik and Tachi, Tomohiro},
year = {2017},
month = {07},
pages = {},
title = {Origamizer: A Practical Algorithm for Folding Any Polyhedron},
doi = {10.4230/LIPIcs.SoCG.2017.34}
}

@article{kilian2017string,
  title={String actuated curved folded surfaces},
  author={Kilian, Martin and Monszpart, Aron and Mitra, Niloy J},
  journal={ACM Transactions on Graphics (TOG)},
  volume={36},
  number={3},
  pages={1--13},
  year={2017},
  publisher={ACM New York, NY, USA}
}

@article{laschi2016soft,
  title={Soft robotics: Technologies and systems pushing the boundaries of robot abilities},
  author={Laschi, Cecilia and Mazzolai, Barbara and Cianchetti, Matteo},
  journal={Science robotics},
  volume={1},
  number={1},
  pages={eaah3690},
  year={2016},
  publisher={American Association for the Advancement of Science}
}

@article{felton2014method,
  title={A method for building self-folding machines},
  author={Felton, Samuel and Tolley, Michael and Demaine, Erik and Rus, Daniela and Wood, Robert},
  journal={Science},
  volume={345},
  number={6197},
  pages={644--646},
  year={2014},
  publisher={American Association for the Advancement of Science}
}

@article{shintake2018soft,
  title={Soft robotic grippers},
  author={Shintake, Jun and Cacucciolo, Vito and Floreano, Dario and Shea, Herbert},
  journal={Advanced materials},
  volume={30},
  number={29},
  pages={1707035},
  year={2018},
  publisher={Wiley Online Library}
}

@article{dudte2016programming,
  title={Programming curvature using origami tessellations},
  author={Dudte, Levi H and Vouga, Etienne and Tachi, Tomohiro and Mahadevan, Lakshminarayanan},
  journal={Nature materials},
  volume={15},
  number={5},
  pages={583--588},
  year={2016},
  publisher={Nature Publishing Group UK London}
}

@inproceedings{yang2024design,
  title={Design and fabrication of string-driven origami robots},
  author={Yang, Peiwen and Li, Shuguang},
  booktitle={2024 IEEE International Conference on Robotics and Automation (ICRA)},
  pages={713--719},
  year={2024},
  organization={IEEE}
}

@article{li2017fluid,
  title={Fluid-driven origami-inspired artificial muscles},
  author={Li, Shuguang and Vogt, Daniel M and Rus, Daniela and Wood, Robert J},
  journal={Proceedings of the National academy of Sciences},
  volume={114},
  number={50},
  pages={13132--13137},
  year={2017},
  publisher={National Academy of Sciences}
}

@article{ppo,
  title={Proximal policy optimization algorithms},
  author={Schulman, John and Wolski, Filip and Dhariwal, Prafulla and Radford, Alec and Klimov, Oleg},
  journal={arXiv preprint arXiv:1707.06347},
  year={2017}
}

@article{double-miura,
  title={Double-Layer Self-Locking Origami Based on Opposite Folding Motion},
  author={Kim, Jae-Kyeong and Ahn, Se Hyeok and Jung, Sun-Pill and Kim, Jemoon and Hwang, Deuk-Gyeom and Kim, Haseon and Kim, Seung-Won and Lee, Dae-Young and Yang, Jinkyu and Cho, Kyu-Jin},
  journal={Soft Robotics},
  pages={21695172251401337},
  year={2025},
  publisher={SAGE Publications Sage CA: Los Angeles, CA}
}

@article{sideline,
  title={Programmable origami strings},
  author={Kamrava, Soroush and Mousanezhad, Davood and Felton, Samuel M and Vaziri, Ashkan},
  journal={Advanced Materials Technologies},
  volume={3},
  number={3},
  pages={1700276},
  year={2018},
  publisher={Wiley Online Library}
}

@article{zhao2021computational,
  title={Computational design methods for cylindrical and axisymmetric waterbomb tessellations},
  author={Zhao, Yan and Li, Shiling and Zhang, Mingyue and Zeng, Lanling and Yang, Yang and Kanamori, Yoshihiro and Mitani, Jun},
  journal={Computer Aided Geometric Design},
  volume={91},
  pages={102037},
  year={2021},
  publisher={Elsevier}
}

@MISC{pybullet,
author =   {Erwin Coumans and Yunfei Bai},
title =    {PyBullet, a Python module for physics simulation for games, robotics and machine learning},
howpublished = {\url{http://pybullet.org}},
year = {2016--2021}
}

@inproceedings{magicball,
  title={A vacuum-driven origami “magic-ball” soft gripper},
  author={Li, Shuguang and Stampfli, John J and Xu, Helen J and Malkin, Elian and Diaz, Evelin Villegas and Rus, Daniela and Wood, Robert J},
  booktitle={2019 International Conference on Robotics and Automation (ICRA)},
  pages={7401--7408},
  year={2019},
  organization={IEEE}
}

@ARTICLE{envelop_soft,
  author={Hao, Yufei and Biswas, Shantonu and Hawkes, Elliot Wright and Wang, Tianmiao and Zhu, Mengjia and Wen, Li and Visell, Yon},
  journal={IEEE Transactions on Robotics}, 
  title={A Multimodal, Enveloping Soft Gripper: Shape Conformation, Bioinspired Adhesion, and Expansion-Driven Suction}, 
  year={2021},
  volume={37},
  number={2},
  pages={350-362},
  doi={10.1109/TRO.2020.3021427}}

@inproceedings{ycb,
  title={The ycb object and model set: Towards common benchmarks for manipulation research},
  author={Calli, Berk and Singh, Arjun and Walsman, Aaron and Srinivasa, Siddhartha and Abbeel, Pieter and Dollar, Aaron M},
  booktitle={2015 international conference on advanced robotics (ICAR)},
  pages={510--517},
  year={2015},
  organization={IEEE}
}

@article{soft_enveloping_gripper,
  title={A soft enveloping gripper with enhanced grasping ability via morphological adaptability},
  author={Hao, Yufei and Wang, Zhongkui and Zhou, Yuzhao and Zhou, Weitai and Cai, Tengfei and Zhang, Jianhua and Sun, Fuchun},
  journal={Advanced Intelligent Systems},
  volume={5},
  number={6},
  pages={2200456},
  year={2023},
  publisher={Wiley Online Library}
}

@article{seo2016theory,
  title={A theory on grasping objects using effectors with curved contact surfaces and its application to whole-arm grasping},
  author={Seo, Jungwon and Yim, Mark and Kumar, Vijay},
  journal={The International Journal of Robotics Research},
  volume={35},
  number={9},
  pages={1080--1102},
  year={2016},
  publisher={SAGE Publications Sage UK: London, England}
}

@inproceedings{honarpardaz2017fast,
  title={Fast grasp tool design: From force to form closure},
  author={Honarpardaz, Mohammadali and Meier, Martin and Haschke, Robert},
  booktitle={2017 13th IEEE Conference on Automation Science and Engineering (CASE)},
  pages={782--788},
  year={2017},
  organization={IEEE}
}

@inproceedings{deimel2017automated,
  title={Automated co-design of soft hand morphology and control strategy for grasping},
  author={Deimel, Raphael and Irmisch, Patrick and Wall, Vincent and Brock, Oliver},
  booktitle={2017 IEEE/RSJ International Conference on Intelligent Robots and Systems (IROS)},
  pages={1213--1218},
  year={2017},
  organization={IEEE}
}

@ARTICLE{hiller2012,
  author={Hiller, Jonathan and Lipson, Hod},
  journal={IEEE Transactions on Robotics}, 
  title={Automatic Design and Manufacture of Soft Robots}, 
  year={2012},
  volume={28},
  number={2},
  pages={457-466},
  doi={10.1109/TRO.2011.2172702}}

@article{
matthews2023,
author = {David Matthews  and Andrew Spielberg  and Daniela Rus  and Sam Kriegman  and Josh Bongard },
title = {Efficient automatic design of robots},
journal = {Proceedings of the National Academy of Sciences},
volume = {120},
number = {41},
pages = {e2305180120},
year = {2023},
doi = {10.1073/pnas.2305180120},
URL = {https://www.pnas.org/doi/abs/10.1073/pnas.2305180120},
eprint = {https://www.pnas.org/doi/pdf/10.1073/pnas.2305180120}}

@INPROCEEDINGS{josh2023,
  author={Pinskier, Josh and Kumar, Prabhat and Langelaar, Matthijs and Howard, David},
  booktitle={2023 IEEE International Conference on Soft Robotics (RoboSoft)}, 
  title={Automated design of pneumatic soft grippers through design-dependent multi-material topology optimization}, 
  year={2023},
  volume={},
  number={},
  pages={1-7},
  doi={10.1109/RoboSoft55895.2023.10122069}}

@article{Wang2024FinBayesAM,
  title={Fin-Bayes: A Multi-Objective Bayesian Optimization Framework for Soft Robotic Fingers},
  author={Xing Wang and Binghao Wang and Josh Pinskier and Yue Xie and James Brett and Richard Scalzo and David Howard},
  journal={Soft Robotics},
  year={2024},
  volume={11},
  pages={791 - 801},
  url={https://api.semanticscholar.org/CorpusID:268512419}
}

@ARTICLE{Sui2022,
  author={Sui, Dongbao and Wang, Tianshuo and Zhao, Sikai and Zhang, Xuehe and Zhao, Jie and Zhu, Yanhe},
  journal={IEEE Robotics and Automation Letters}, 
  title={An Enveloping Soft Gripper With High-Load Carrying Capacity: Design, Characterization and Application}, 
  year={2022},
  volume={7},
  number={1},
  pages={373-380},
  doi={10.1109/LRA.2021.3126907}}

@ARTICLE{Chen2023origami,
  author={Chen, Wei-Hsi and Yang, Woohyeok and Peach, Lucien and Koditschek, Daniel E. and Sung, Cynthia R.},
  journal={IEEE Transactions on Robotics}, 
  title={Kinegami: Algorithmic Design of Compliant Kinematic Chains From Tubular Origami}, 
  year={2023},
  volume={39},
  number={2},
  pages={1260-1280},
  doi={10.1109/TRO.2022.3206711}}

@INPROCEEDINGS{Chen2020hopper,
  author={Chen, Wei-Hsi and Misra, Shivangi and Caporale, J. Diego and Koditschek, Daniel E. and Yang, Shu and Sung, Cynthia R.},
  booktitle={2020 3rd IEEE International Conference on Soft Robotics (RoboSoft)}, 
  title={A Tendon-Driven Origami Hopper Triggered by Proprioceptive Contact Detection}, 
  year={2020},
  volume={},
  number={},
  pages={373-380},
  doi={10.1109/RoboSoft48309.2020.9116040}}

@INPROCEEDINGS{Feshbach2023,
  author={Feshbach, Daniel and Wu, Xuelin and Vasireddy, Satviki and Beardell, Louis and To, Bao and Baryshnikov, Yuliy and Sung, Cynthia},
  booktitle={2023 IEEE/RSJ International Conference on Intelligent Robots and Systems (IROS)}, 
  title={CurveQuad: A Centimeter-Scale Origami Quadruped that Leverages Curved Creases to Self-Fold and Crawl with One Motor}, 
  year={2023},
  volume={},
  number={},
  pages={2485-2492},
  doi={10.1109/IROS55552.2023.10342339}}

@article{huang2026crabedm,
  author  = {Huang, Xia and Xiang, Haomin and Wu, Chang and Xu, Jiabin and Ding, Jun and Peng, Yanhong and Funabora, Yuki},
  title   = {{CRAB-EDM}: A Multi-Modal Underwater Crab-Inspired Robot With Temporally Sequenced Electro-Discharge Modulation},
  journal = {IEEE Robotics and Automation Letters},
  volume  = {11},
  number  = {4},
  pages   = {3939--3946},
  year    = {2026},
  doi     = {10.1109/LRA.2026.3663817}
}

\end{document}